\documentclass[letterpaper]{article} 
\usepackage{aaai25}  
\usepackage{times}  
\usepackage{helvet}  
\usepackage{courier}  
\usepackage[hyphens]{url}  
\usepackage{graphicx} 
\usepackage{natbib}  
\usepackage{caption} 
\usepackage{algorithm}
\usepackage{algorithmic}
\usepackage[caption=false,font=footnotesize,labelfont=rm,textfont=rm]{subfig}
\usepackage{multirow}
\usepackage{threeparttable}
\usepackage{booktabs}
\usepackage{pifont}
\usepackage{amsmath}
\usepackage{amssymb}
\usepackage{newfloat}
\usepackage{listings}
\DeclareCaptionStyle{ruled}{labelfont=normalfont,labelsep=colon,strut=off} 
\floatstyle{ruled}
\newfloat{listing}{tb}{lst}{}
\floatname{listing}{Listing}
\title{Map2Traj: Street Map Piloted Zero-shot Trajectory Generation \\ with Diffusion Model}
\author {
        Zhenyu~Tao\textsuperscript{\rm 1,2},
    Wei~Xu\textsuperscript{\rm 1,2},
   and Xiaohu~You\textsuperscript{\rm 1,2}
}
\affiliations {
    \textsuperscript{\rm 1} National Mobile Communications Research Lab, Southeast University, China\\
    \textsuperscript{\rm 2} Pervasive Communication Research Center, Purple Mountain Laboratories, China\\
    \{zhenyu\_tao, wxu, xhyu\}@seu.edu.cn
}

\usepackage{bibentry}

\begin{document}

\maketitle

\begin{abstract}

User mobility modeling serves a crucial role in analysis and optimization of contemporary wireless networks. Typical stochastic mobility models, e.g., random waypoint model and Gauss Markov model, can hardly capture the distribution characteristics of users within real-world areas. State-of-the-art trace-based mobility models and existing learning-based trajectory generation methods, however, are frequently constrained by the inaccessibility of substantial real trajectories due to privacy concerns. In this paper, we harness the intrinsic correlation between street maps and trajectories and develop a novel zero-shot trajectory generation method, named Map2Traj, by exploiting the diffusion model. We incorporate street maps as a condition to consistently pilot the denoising process and train our model on diverse sets of real trajectories from various regions in Xi'an, China, and their corresponding street maps. With solely the street map of an unobserved area, Map2Traj generates synthetic trajectories that not only closely resemble the real-world mobility pattern but also offer comparable efficacy. Extensive experiments validate the efficacy of our proposed method on zero-shot trajectory generation tasks in terms of both trajectory and distribution similarities. In addition, a case study of employing Map2Traj in wireless network optimization is presented to validate its efficacy for downstream applications.

\end{abstract}

%

\section{Introduction} 





With the long-term evolution of cellular networks in terms of heterogeneity, density, and multi-band usage, the accuracy of user mobility modeling has become increasingly crucial for performance evaluation and optimization of wireless communication networks. In the realm of learning-based network optimization, involving resource management \cite{9329087}, user association \cite{10298039}, and edge computing \cite{10024766}, user mobility models stand as the cornerstone for constructing training environments and digital twins \cite{tao2023wireless} for artificial intelligence (AI) models like deep reinforcement learning (DRL) agents. 

Most existing network optimization studies employed random mobility models, typically the random waypoint model  \cite{Johnson1996} and the Gauss Markov model \cite{752157}, to represent user mobility patterns. While these models can partially simulate user movement, their direct adherence to random probability distributions causes a significant mismatch in the spatial distribution of users compared to real-world scenarios. This mismatch can lead to significant performance degradation when deploying these models in practice. Although this issue can be alleviated by incorporating real trace-based mobility models to some extent, user trajectories are often unfortunately inaccessible due to data acquisition costs and privacy concerns \cite{8673556}. 

The urgent need for generating high-fidelity synthetic trajectories as alternatives to real ones has driven the exploration of learning-based trajectory generation methods. Over recent years, AI models such as generative adversarial network (GAN) \cite{goodfellow2020generative}, sequence-to-sequence (Seq2Seq) \cite{sutskever2014sequence}, and diffusion model \cite{NEURIPS2020_4c5bcfec} have been applied to trajectory generation with promising results \cite{9338370,trajgen,tstrajgen,SynMob,DiffTraj}. However, these methods typically require a substantial number of real trajectories to learn the specific trajectory distribution within an area, creating a \textbf{paradox}. That is, AI models struggle to generate realistic and useful trajectories without ample real trajectories, yet when real data becomes sufficient to create trace-based mobility models, the AI models tend to be redundant.

\begin{figure}[!t]
\centering

\subfloat[Street map]{\includegraphics[width=0.16\textwidth]{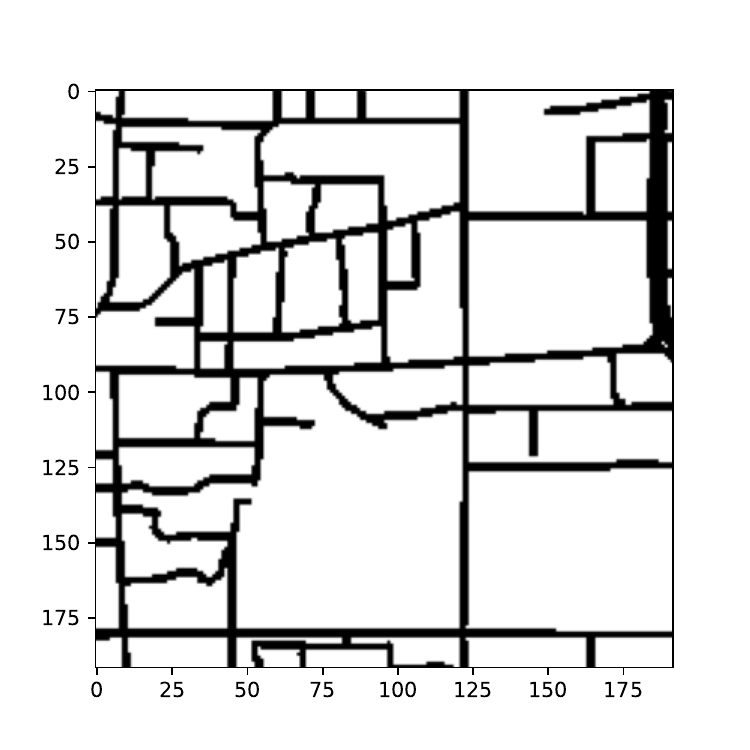}\label{Fig1.1}}
\subfloat[Real trajectories]{\includegraphics[width=0.16\textwidth]{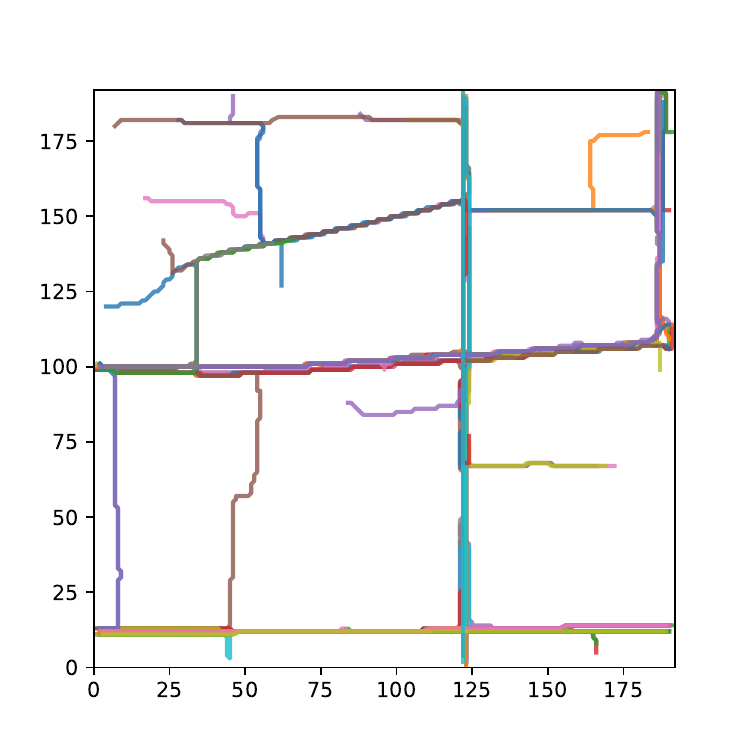}\label{Fig1.2}}
\subfloat[Real heatmap]{\includegraphics[width=0.16\textwidth]{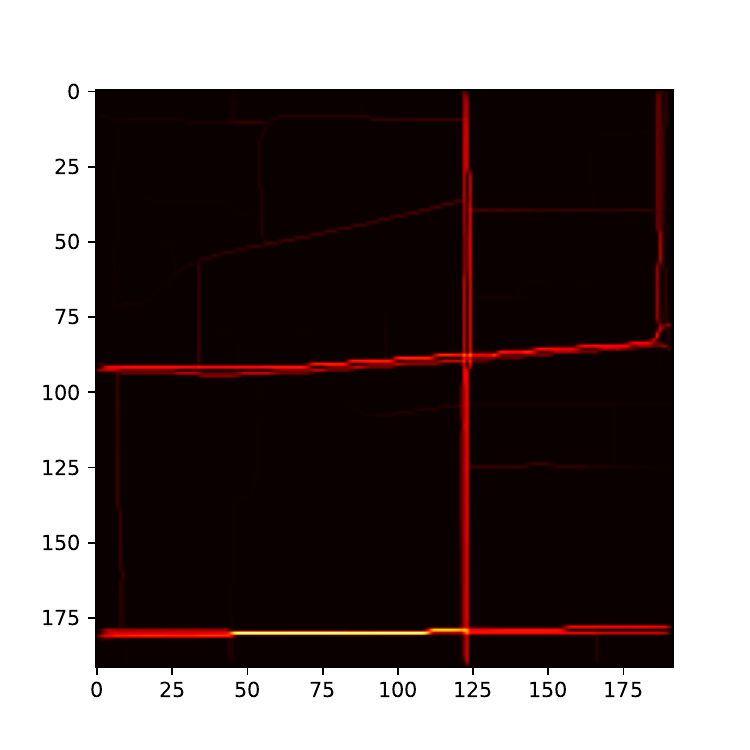}\label{Fig1.3}}

\caption{Street map, trajectories, and trajectory heatmap}
\label{Fig1}
\end{figure}

Inspired by zero-shot image generation \cite{pmlr-v139-ramesh21a}, which enables the creation of images from descriptions unseen during training, we try to devise a similar approach for trajectory generation. This method, termed \textbf{zero-shot trajectory generation}, aims to generate realistic user trajectories for unobserved areas. The question arises: Can we apply zero-shot generation techniques to trajectory generation? More specifically, is there data that is both readily accessible and closely related to real trajectories, akin to the relationship between text and the images? The answer lies in street maps, which are usually open-source and available on platforms like OpenStreetMap\footnote{https://www.openstreetmap.org/}. Street maps exhibit a strong correlation with user trajectories as illustrated in Fig.~\ref{Fig1}, which respectively show the street map for a specific area, 100 trajectories in this area, each assigned a color, and a heatmap of the trajectory distribution. This correlation forms the foundation of our proposed methodology, which uses street maps as pilots in trajectory generation.

Building on these motivations and the observed correlation, we propose a street-\textbf{Map}-pilot zero-shot \textbf{Traj}ectory generation (Map2Traj) method. With solely a street map input, Map2Traj generates synthetic trajectories similar to real ones in both trajectory and spatial distributions. Moreover, the synthetic trajectories generated by Map2Traj retain similar efficacy to real ones, enabling learning-based methods trained in a Map2Traj-based simulation environment to be deployed in real scenarios with minimal performance degradation due to mobility models. To summarize, the contributions of this work are as follows.

\begin{itemize}
\item{We develop the Map2Traj method for zero-shot trajectory generation, leveraging the diffusion model to simulate real-world trajectories based on street map. To the best of our knowledge, this is the first work to achieve zero-shot trajectory generation. 
}
\item{We validate the efficacy of Map2Traj through comprehensive experiments, proving that our method can generate high-fidelity trajectories for areas beyond the training set, with considerable similarity to real trajectories.
}
\item{We examine the efficacy of Map2Traj in a network optimization task, specifically user association and load balancing in a multi-cell and multiuser wireless communication network. The results indicate that the Map2Traj-based mobility model significantly outperforms traditional random mobility models and performs nearly as well as the model using actual data.
}

\end{itemize}

\section{Related Work} 

Before introducing our proposed method, we review related works on trajectory generation and analyze the limitations that prevent these methods from achieving zero-shot generation capabilities.

Initially, trajectory generation methods were developed for synthesizing mobility data to safeguard the privacy of data providers. \citet{liu2018trajgans} first proposed to use GANs for trajectory generation, albeit without providing a detailed approach. TrajGAIL \cite{9338370} employed generative adversarial imitation learning (GAIL), combining DRL and GAN to generate trajectories through a series of next-location predictions. TrajGen \cite{trajgen} transformed trajectories into images and used a deep convolutional GAN (DCGAN) to generate virtual trajectory images. TS-TrajGen \cite{tstrajgen} integrated GAN with the mobility analysis method, including the A* algorithm and mobility yaw reward, to enhance the model performance. DiffTraj (SynMob) \cite{SynMob,DiffTraj} applied a diffusion model to generate synthetic trajectories while preserving spatial-temporal features extracted from real trajectories. These studies have demonstrated commendable performance in generating privacy-preserving synthetic trajectories.

However, these methods fall short when it comes to zero-shot trajectory generation for unobserved new areas. Specifically, TrajGAIL simply samples actions from generated action probability distribution and constructs trajectory autoregressively, without the capacity to introduce data from new areas. Although TrajGen uses street map data to filter and calibrate generated trajectories through map matching \cite{mapmatching}, the generated trajectories adhere to the training set distribution, rather than that in new areas. Building further upon TrajGen, TS-TrajGen utilizes street maps to select and construct the best continuous trajectory with the A* algorithm. This process, however, remains confined to trajectories that adhere to the original distribution. Alternatively in a conditional generation manner, DiffTraj employs the diffusion model and incorporates prior knowledge of trip data, such as the travel time, average speed, and distance. While these complementary knowledge do improve generation performance, they do not enable the transfer of trajectory generation to new areas.

In contrast to these approaches, our Map2Traj method integrates the street map with complete information into the trajectory generation process via a diffusion model. Our training set encompasses a diverse range of trajectories and corresponding street maps from various areas, instead of area-specific trajectories, allowing the model to learn the intrinsic relationship between street maps and trajectories. These innovations endow our model with the unique capability for zero-shot trajectory generation. A comparative analysis of our method against existing works is detailed in Table~\ref{tab1}.

\begin{table*}[]
    \centering
    \begin{tabular}{c|c|c|c|c}
        \toprule
         Method & Model & Training data & Inference data & Zero-shot generation\\
        \midrule
        TrajGAIL (ICDM, 2020) & GAIL & Trajectories & Random sampling & \ding{55}\\
        TrajGen (KDD, 2021) & DCGAN & Trajectories + maps& Noise + maps & \ding{55}\\
        TS-TrajGen (AAAI, 2023) & GAN & Trajectories + maps & Noise + maps & \ding{55}\\
        DiffTraj (NIPS, 2023) & Diffusion model & Trajectories + trip data & Noise  + trip data & \ding{55}\\
        Map2Traj (Proposed) & Diffusion model & Trajectories + maps & Noise + maps & \ding{51}\\
        \bottomrule
    \end{tabular}
    \caption{Comparison of trajectory generation methods}
    \label{tab1}
\end{table*}

\section{Preliminary} 
In this section, we introduce the definitions and notations used in this paper.

\textbf{Trajectory:} A trajectory is typically a sequence of location points consisting of latitude and longitude. For this paper focusing on wireless network optimization, trajectories are restricted to a relatively small-scale urban area, specifically a 1.92 km $\times$ 1.92 km square. We define a trajectory as a series of relative coordinates, denoted by $\boldsymbol{l} = \{c_1, c_2, \dots, c_n\}$, where each $c_i$ is a coordinate $\left[x_i, y_i\right]$ with $\{x_i, y_i\} \in \left[0,1920\right]$. 
Considering the spatial consistency of wireless channels, we discretize the coordinates $\left[x_i, y_i\right]$ to integer multiples of 10, focusing on macroscopic position changes. This allows us to easily transform the trajectory sequence into a 192$\times$192 binary image for further processing. The transformation method between trajectory and image is well-documented in literature like \cite{endo2016classifying}, and it has been used in TrajGen \cite{trajgen}.

\textbf{Street Map:} A street map is conventionally denoted as a graph, where edges correspond to road segments and nodes to road junctions. To align with the trajectories, we also convert each street map into an image of the same 192$\times$192 dimension, denoted by $\boldsymbol{m}$.

\textbf{Problem Statement:} The training set includes a set of street maps $\mathcal{M} = \{\boldsymbol{m}^1,\boldsymbol{m}^2,\dots,\boldsymbol{m}^K\}$ and corresponding sets of real-world trajectories $\mathcal{T} =\{\mathcal{L}^1,\mathcal{L}^2,\dots,\mathcal{L}^K\}$. Each $\mathcal{L}^i = \{\boldsymbol{l}^{i,1},\boldsymbol{l}^{i,2},\dots,\boldsymbol{l}^{i,J}\}$ is a set of $J$ real trajectories within the area of street map $\boldsymbol{m}^i$. The objective of zero-shot trajectory generation is to develop a generative model trained on this training set. For an unobserved street map $\boldsymbol{m}^o \notin \mathcal{M}$, this model should be capable of generating synthetic trajectories that: 1) closely resemble real trajectories; 2) exhibit a spatial distribution akin to that of the real trajectory set; and 3) have an efficacy close to that of real trajectories in downstream applications and analyses.

\section{The Map2Traj Approach}

\begin{figure}[!t]
\centering
\includegraphics[width=0.5\textwidth]{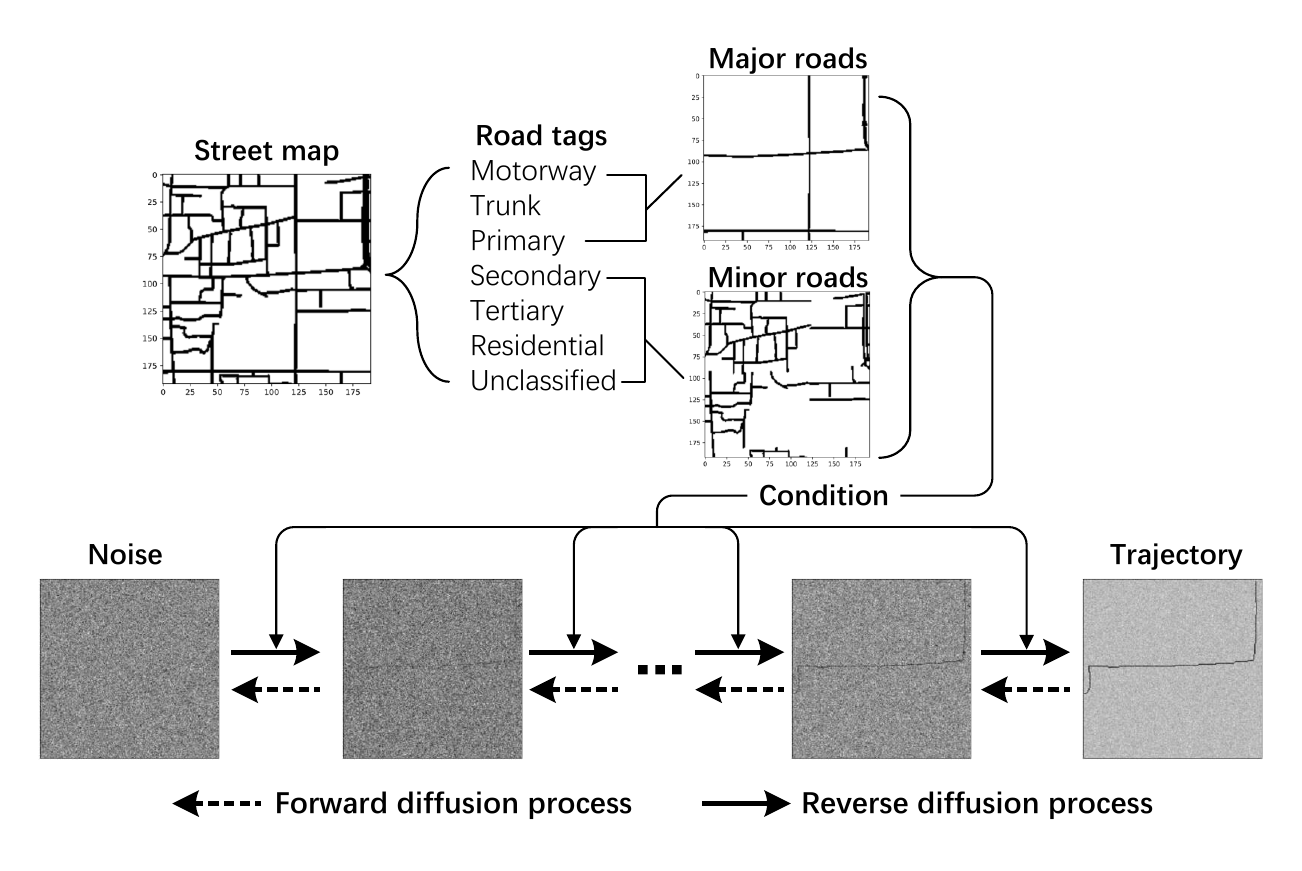}
\caption{Map2Traj framework.}
\label{Fig2}
\end{figure}

The key to zero-shot learning is to associate observed and unobserved objects through some form of auxiliary information, which encodes the inherent properties of objects \cite{xian2017zero}. In our study, the objects are trajectories following different area-specific distributions, while the auxiliary information is the street map. By understanding the relationship between trajectories and maps through extensive training data, Map2Traj generates synthetic trajectories for unobserved areas piloted by their street maps, that is, zero-shot trajectory generation.

In particular, our Map2Traj approach is based on the diffusion model, which consists of a forward diffusion process and a reverse diffusion process (denoising) for generation. The basic diffusion model converts samples from a standard Gaussian distribution into outputs that follow a single target distribution by iterative denoising. By integrating trajectories from various areas and utilizing relevant street maps as conditional inputs, Map2Traj extends the original single target distribution into multiple area-specific target distributions corresponding to given street maps. As a result, Map2Traj can estimate the trajectory distribution through an unobserved street map and generate synthetic trajectories that conform to this distribution through sampling. The entire process is illustrated in Fig. \ref{Fig2}. Both the trajectory $\boldsymbol{l}$ and street map $\boldsymbol{m}$ in Map2Traj are represented as images.

\subsection{Forward Diffusion Process}

The forward diffusion process in Map2Traj is a Markovian process that iteratively adds Gaussian noise $\mathcal{N}(\cdot)$ to a trajectory data $\boldsymbol{l}_0 \equiv \boldsymbol{l}$ over $T$ time steps:

\begin{align}
& q\left(\boldsymbol{l}_{t+1} \mid \boldsymbol{l}_t\right)=\mathcal{N}\left(\boldsymbol{l}_{t+1} ;  \sqrt{\alpha_t} \boldsymbol{l}_{t},\left(1-\alpha_t\right) \mathbf{I}\right), \\
& q\left(\boldsymbol{l}_{1: T} \mid \boldsymbol{l}_0\right)=\prod_{t=1}^T q\left(\boldsymbol{l}_t \mid \boldsymbol{l}_{t-1}\right),
\end{align}
where $\alpha_t \text{ for } t=1,2,\dots,T$ are hyper-parameters of the noise schedule, and $\mathcal{N}(x;\mu,\sigma)$ represents the normal distribution of mean $\mu$ and covariance $\sigma$ that produces $x$. The forward process with $\alpha_t$ is constructed to make $\boldsymbol{l}_T$ virtually indistinguishable from Gaussian noise at the $T$-th step. The forward process at the $t$-th step can also be marginalized as follows:
\begin{align}
q\left(\boldsymbol{l}_t \mid \boldsymbol{l}_0\right)=\mathcal{N}\left(\boldsymbol{l}_t ;  \sqrt{\gamma_t} \boldsymbol{l}_0,\left(1-\gamma_t\right) \mathbf{I}\right), \label{eq3}
\end{align}
where $\gamma_t=\prod_{i=1}^t \alpha_i$.
Additionally, the parameterization of the Gaussian distribution of the forward process allows a closed-form formulation of the posterior distribution of $\boldsymbol{l}_{t-1}$ given $\left(\boldsymbol{l}_0, \boldsymbol{l}_t\right)$. It follows
\begin{align}
q\left(\boldsymbol{l}_{t-1} \mid \boldsymbol{l}_0, \boldsymbol{l}_t\right)=\mathcal{N}\left(\boldsymbol{l}_{t-1} ; \boldsymbol{\mu}, \sigma^2 \mathbf{I}\right), \label{eq4}
\end{align}
where $\boldsymbol{\mu}=\frac{\sqrt{\gamma_{t-1}}\left(1-\alpha_t\right)}{1-\gamma_t} \boldsymbol{l}_0+\frac{\sqrt{\alpha_t}\left(1-\gamma_{t-1}\right)}{1-\gamma_t} \boldsymbol{l}_t$ and $\sigma^2=\frac{\left(1-\gamma_{t-1}\right)\left(1-\alpha_t\right)}{1-\gamma_t}$.

\subsection{Reverse Diffusion Process}
In Map2Traj, the reverse diffusion process, also known as the denoising process, is formulated as follows:
\begin{equation}
    p_{\boldsymbol{\theta}}\left(\boldsymbol{l}_{0: T}\mid \boldsymbol{m}\right)=p\left(\boldsymbol{l}_T\right) \prod_{t=1}^T p_{\boldsymbol{\theta}}\left(\boldsymbol{l}_{t-1} \mid \boldsymbol{l}_t , \boldsymbol{m}\right), \label{denoise}
\end{equation}
where $p(\boldsymbol{l}_T) = \mathcal{N}(\boldsymbol{l}_T;0, \mathbf{I})$. Map2Traj undergoes training and performs inference through the reverse diffusion process.
\subsubsection{Training}
Given a noisy trajectory $\boldsymbol{l}_t$ sampling from Eq.~(\ref{eq3}), we have
\begin{align}
\boldsymbol{l}_t=\sqrt{\gamma_t} \boldsymbol{l}_0+\sqrt{1-\gamma_t} \boldsymbol{\epsilon},\ \ \boldsymbol{\epsilon} \sim \mathcal{N}(\mathbf{0}, \mathbf{I})\label{eq5}
\end{align}
where the goal is to recover the target trajectory $\boldsymbol{l}_0$. Our neural network model is parameterized by $f_{\boldsymbol{\theta}}(\boldsymbol{m}, \boldsymbol{l}_t, t)$, conditioned on the street map $\boldsymbol{m}$, a noisy trajectory $\boldsymbol{l}_t$, and the noise level indicated by the time step $t$. Training of Map2Traj involves predicting the noise vector $\boldsymbol{\epsilon}$ by minimizing the mean squared error loss. That is,
\begin{align}
\min _{\boldsymbol{\theta}} \mathbb{E}_{\boldsymbol{m},\boldsymbol{l},t,\boldsymbol{\epsilon}}\|f_{\boldsymbol{\theta}}(\boldsymbol{m}, \underbrace{\sqrt{\gamma_t} \boldsymbol{l}_0+\sqrt{1-\gamma_t} \boldsymbol{\epsilon}}_{\boldsymbol{l}_t}, t)-\boldsymbol{\epsilon}\|^2.
\end{align}

\subsubsection{Inference}
The sampling process of the diffusion model starts at pure Gaussian noise $\boldsymbol{l}_T$, followed by $T$ refinement steps. Given any noisy trajectory $\boldsymbol{l}_t$, we can approximate the target trajectory by rearranging the terms in Eq.~(\ref{eq5}) as
\begin{equation}
    \hat{\boldsymbol{l}}_0=\frac{1}{\sqrt{\gamma_t}}\left(\boldsymbol{l}_t-\sqrt{1-\gamma_t} f_{\boldsymbol{\theta}}\left(\boldsymbol{m}, \boldsymbol{l}_t, t\right)\right).
\end{equation}
Substituting estimate $\hat{\boldsymbol{l}}_0$ into Eq.~(\ref{eq4}), we parameterize the mean of $p_{\boldsymbol{\theta}}\left(\boldsymbol{l}_{t-1} \mid \boldsymbol{l}_t , \boldsymbol{m}\right)$ in Eq.~(\ref{denoise}) as
\begin{equation}
    \mu_{\boldsymbol{\theta}}\left(\boldsymbol{m}, \boldsymbol{l}_t, t\right)=\frac{1}{\sqrt{\alpha_t}}\left(\boldsymbol{l}_t-\frac{1-\alpha_t}{\sqrt{1-\gamma_t}} f_{\boldsymbol{\theta}}(\boldsymbol{m}, \boldsymbol{l}_t, t)\right).
\end{equation}
And the variance of $p_{\boldsymbol{\theta}}\left(\boldsymbol{l}_{t-1} \mid \boldsymbol{l}_t , \boldsymbol{m}\right)$ is approximated as $(1-\alpha_t)$, following the setting in \cite{NEURIPS2020_4c5bcfec}. With this parameterization, the sampling can be executed iteratively as follows:
\begin{equation}
\boldsymbol{l}_{t-1} \leftarrow \frac{1}{\sqrt{\alpha_t}}\left(\boldsymbol{l}_t-\frac{1-\alpha_t}{\sqrt{1-\gamma_t}} f_{\boldsymbol{\theta}}\left(\boldsymbol{m}, \boldsymbol{l}_t, t\right)\right)+\sqrt{1-\alpha_t} \boldsymbol{\epsilon},
\end{equation}
where $\boldsymbol{\epsilon} \sim \mathcal{N}(\mathbf{0}, \mathbf{I})$.

\subsection{Architecture of Map2Traj}

The architecture of Map2Traj is based on a 192$\times$192 U-Net model \cite{ronneberger2015u}, with multiple modifications to improve its performance such as attention blocks \cite{oktay2018attention} and group normalization \cite{Wu_2018_ECCV}. A distinctive feature of Map2Traj is the incorporation of street map data through concatenation, which allows the model to be conditioned on the spatial information inherent in the maps.

\subsection{Other Technical Details}
We also employ a few other techniques to further enhance the performance of Map2Traj.

\subsubsection{Street Map Splitting}
While a single binary image can effectively convey the spatial layout of a street map, it falls short in depicting the distinct characteristics of various road types. In the OpenStreetMap dataset, roads are tagged with attributes such as \textit{Motorway}, \textit{Primary}, and \textit{Residential}. To exploit these attributes, we categorize roads into multiple groups, create binary images for each group, and merge these into a multi-channel binary image. Due to the computational complexity of the diffusion model, we simplify this process by dividing the roads into two groups: major and minor roads, as illustrated in Fig.~\ref{Fig2}.

\subsubsection{Data Augmentation}
We notice that the correlation between street map and trajectory, as shown in Fig.~\ref{Fig1}, remains consistent under transformations such as rotation and reflection. This inherent property can be leveraged for data augmentation during training. We randomly rotate and flip both street maps and trajectory data to enhance the generalization capability of Map2Traj.






\section{Experiments}
In this section, we evaluate the efficacy of Map2Traj in the zero-shot trajectory task by comparing the fidelity of generated trajectories with real ones. In addition, we employ Map2Traj in a popular task of wireless network optimization to validate its efficacy in practice.

\subsection{Dataset Description} Map2Traj is trained using a real-world trajectory dataset from Xi'an, China, recorded in 2016 \cite{didi2017gaia}, alongside the OpenStreetMap dataset of the same year. These datasets are sourced from the ChinaGEOSS Data Sharing Network\footnote{https://chinageoss.cn/}. The trajectories are all within the latitude range of 34.21 to 34.28 and the longitude range of 108.912 to 108.996. To avoid data leakage in the zero-shot generation, the training set is limited to longitudes between 108.912 and 108.974, while the test set extends from 108.974 to 108.996.

\subsection{Evaluation Metrics}
We employ a suite of metrics to evaluate the quality of generated trajectories, in terms of on both trajectory and distribution similarities.
\subsubsection{Trajectory Similarity} 
\begin{itemize}
    \item \textbf{Edit Distance on Real Sequences (EDR):} EDR \cite{editdistance} quantifies the minimum number of operations required to make two trajectories match. A match is defined when the distance between corresponding points is less than a threshold of $\tau = 20$ meters.

    \item \textbf{Dynamic Time Wrapping (DTW):} DTW \cite{dtw} calculates the squared Euclidean distance between two trajectories through a dynamic programming alignment algorithm.
\end{itemize}
Both metrics are widely used in mobility analysis \cite{gis}.

\subsubsection{Distribution Similarity} 
\begin{itemize}
    \item \textbf{Cosine Similarity:} Cosine similarity is a widely used measure of similarity between two vectors. While it reflects the similarity between probability distributions, it falls short in expressing the spatial correlation between adjacent blocks in two-dimensional (2D) distributions.

    \item \textbf{Wasserstein Distance:} To address the limitations of cosine similarity, we introduce the Wasserstein distance \cite{ruschendorf1985wasserstein}, which is defined as the cost of the optimal transport plan for moving the mass in the predicted measure to match that in the target. In this context, it measures the effort required to transform the spatial distribution of generated trajectories into that of real trajectories. However, computing Wasserstein distance for 2D distributions entails solving a complex high-dimensional linear programming problem, which is computationally impractical for 192$\times$192 arrays. Consequently, we use sliced Wasserstein distance as an alternative \cite{kolouri2019generalized}.
\end{itemize}

\subsection{Baseline Methods}
Due to the limitation of existing learning-based methods for zero-shot trajectory generation, we primarily compare Map2Traj with traditional random mobility models.

One of the most commonly used models in wireless networks is the random waypoint model (RWP), where the trajectory is formed by constantly moving to a randomly chosen destination and then selecting the next one arbitrarily. To adapt this model to geographical constraints, we develop a variant, termed map-restricted random waypoint (M-RWP), where the destinations are confined within the street map area. The trajectory between points is determined using a breadth-first search (BFS) algorithm to ensure that the shortest path remains within streets.

Additionally, we considered the Gauss Markov model (GM), characterized by using a stochastic process to model changes in user velocity and direction. Similarly, we introduce the map-restricted Gauss Markov model (M-GM) to restrict user movements within street areas.

Despite its lack of zero-shot trajectory generation capability, we include the state-of-the-art trajectory generation model DiffTraj \cite{DiffTraj}, to benchmark the generation quality of our proposed model. It is important to note that for this comparison, DiffTraj was trained on the complete trajectory dataset \cite{didi2017gaia}, including those from the test area. The relevant data is sourced from the DiffTraj-generated synthetic dataset, SynMob, provided by the authors of DiffTraj in \cite{SynMob}.


\subsection{Generation Performance}
\begin{figure}[t!]
\centering
\subfloat[Real street map]{\includegraphics[width=0.155\textwidth]{Fig/StreetMap_black.pdf}\label{Fig1.1}}
\subfloat[Real trajectories]{\includegraphics[width=0.155\textwidth]{Fig/real_traj.pdf}\label{Fig1.2}}
\subfloat[Real heatmap]{\includegraphics[width=0.155\textwidth]{Fig/real_hot.pdf}\label{Fig1.3}}

\subfloat[RWP]{\includegraphics[width=0.155\textwidth]{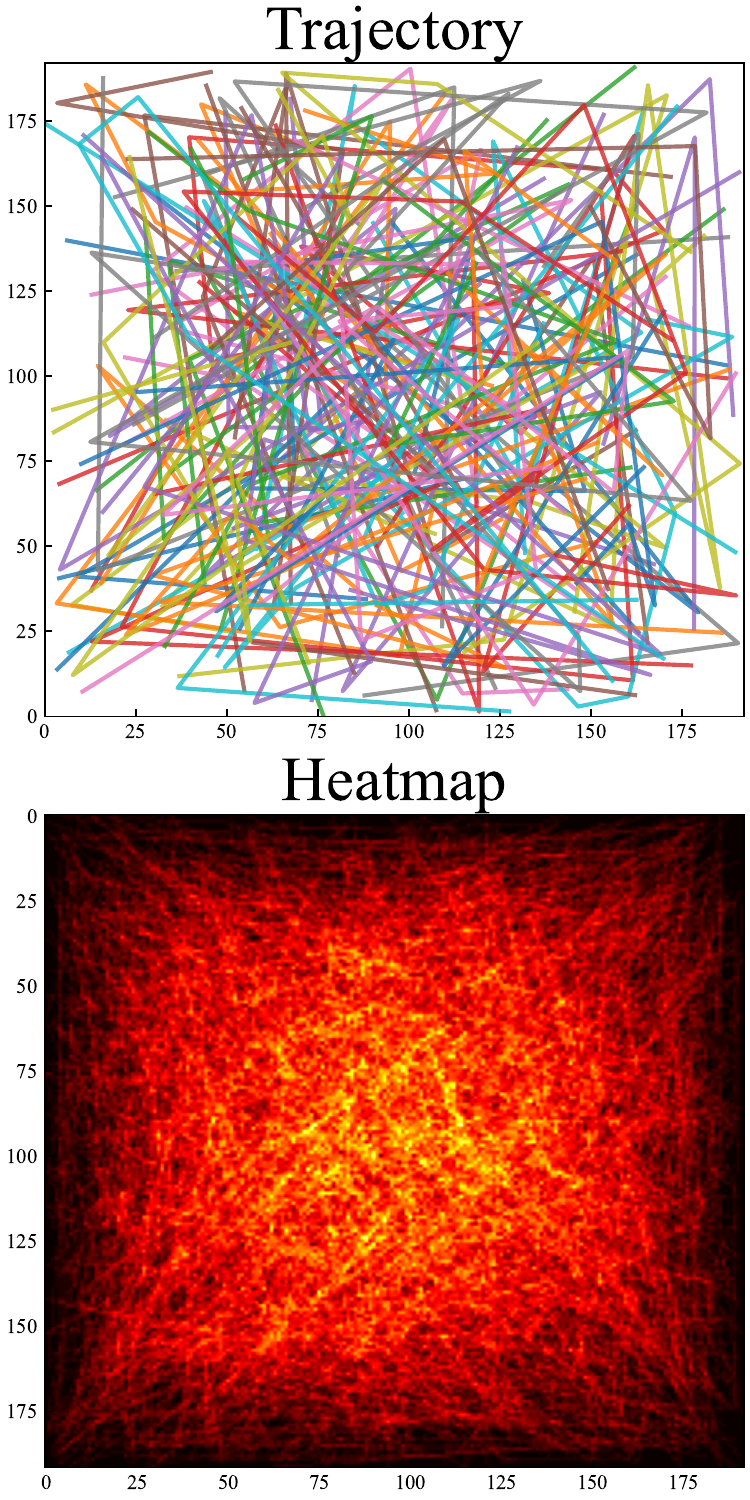}}
\subfloat[GM]{\includegraphics[width=0.155\textwidth]{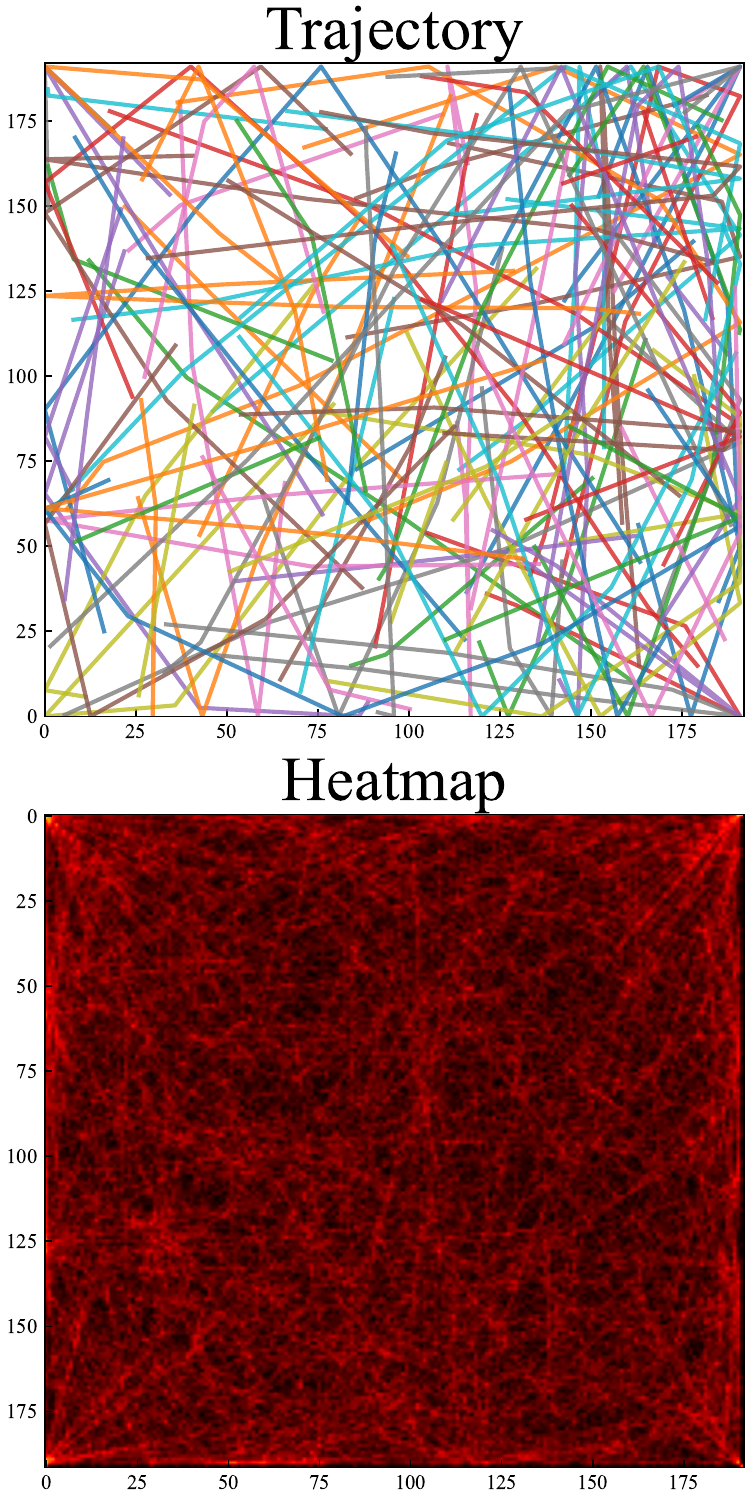}}
\subfloat[M-RWP]{\includegraphics[width=0.155\textwidth]{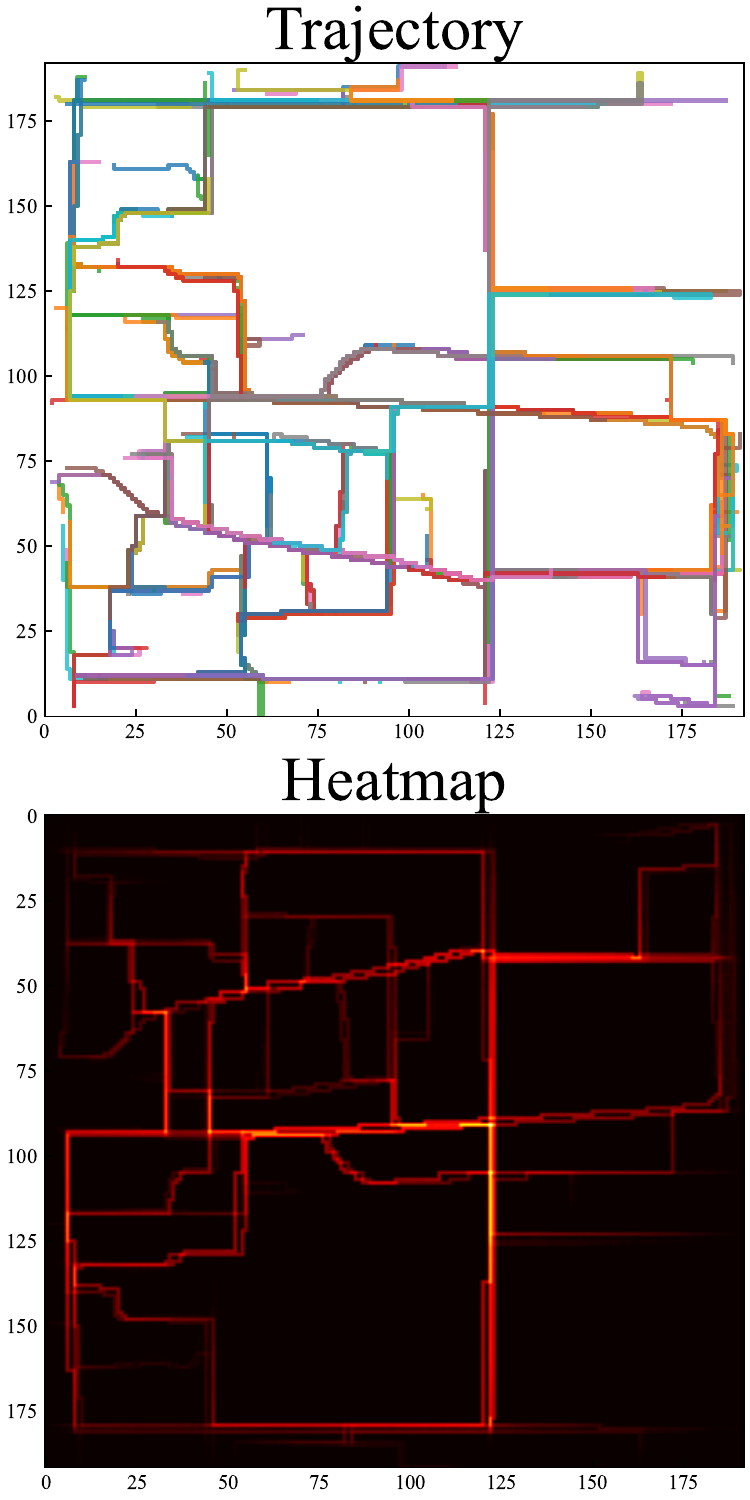}}

\subfloat[M-GM]{\includegraphics[width=0.155\textwidth]{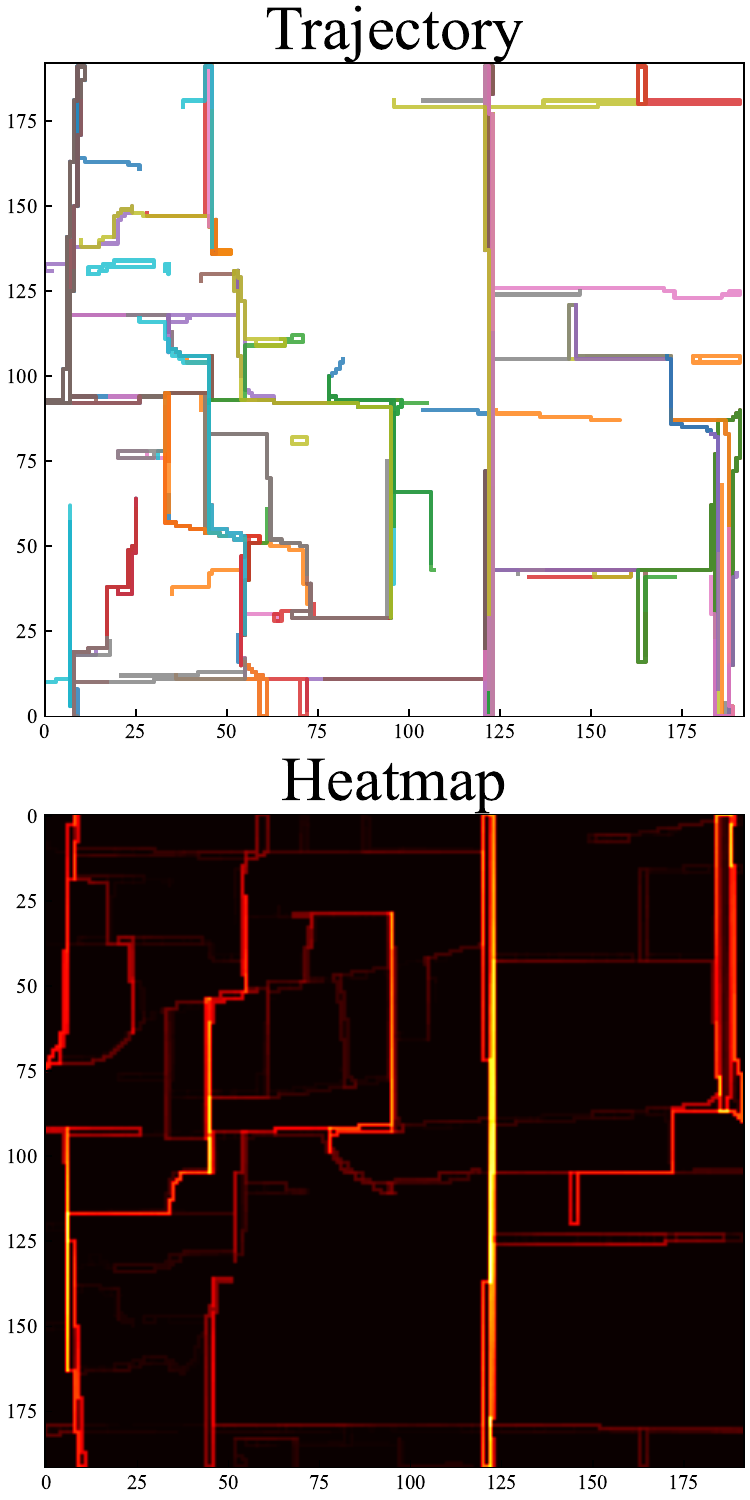}}
\subfloat[DiffTraj]{\includegraphics[width=0.155\textwidth]{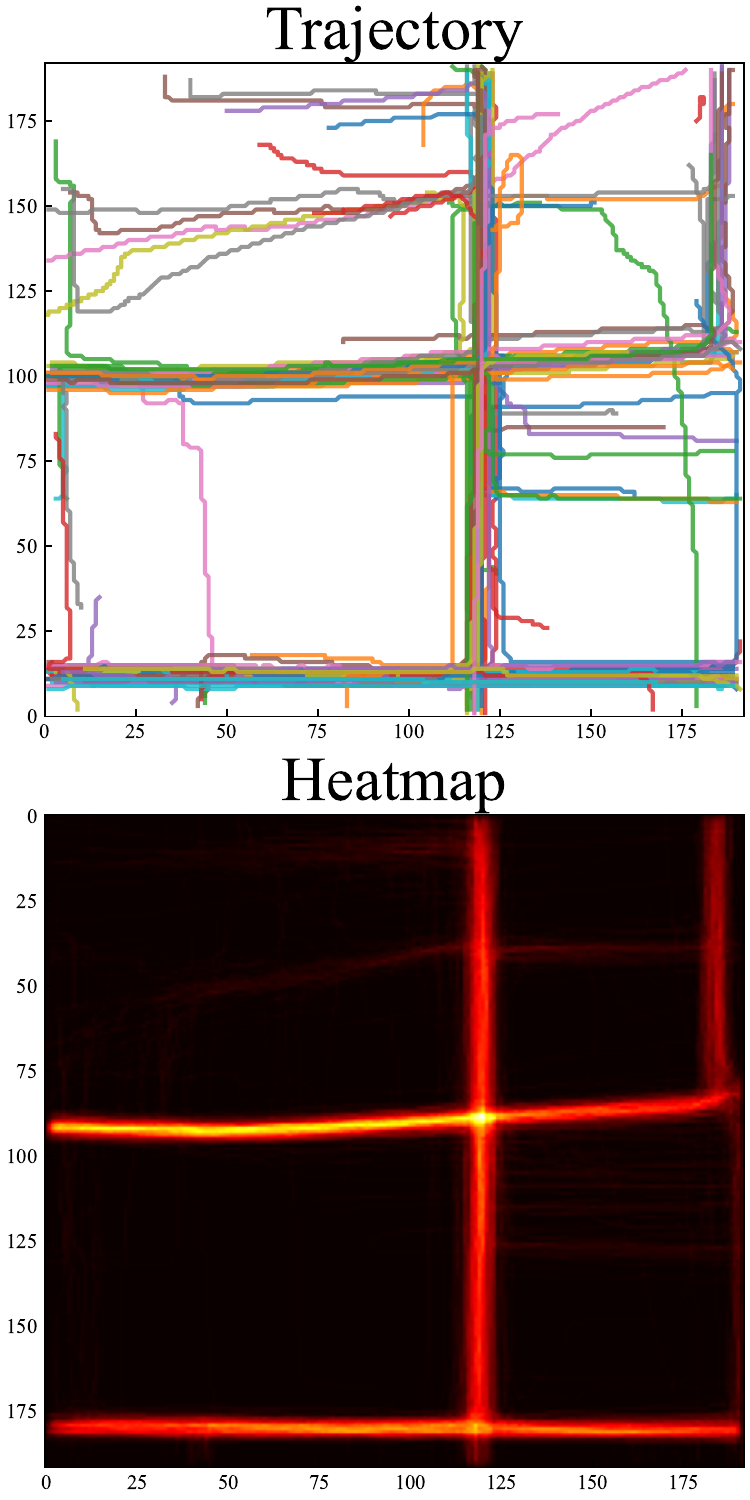}}
\subfloat[Map2Traj]{\includegraphics[width=0.155\textwidth]{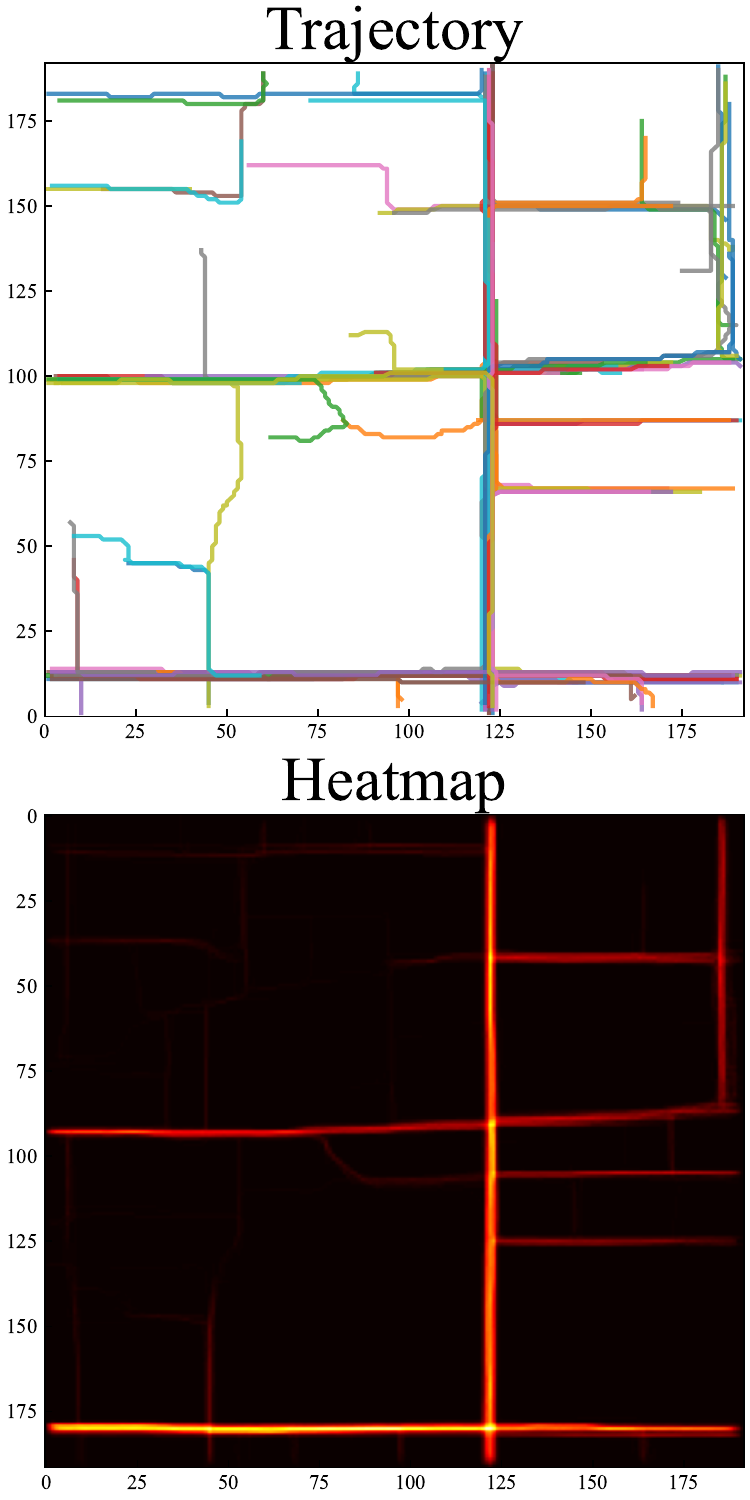}}

\caption{Generated trajectories and heatmaps.} 
\label{Fig3}
\end{figure}

We select the area depicted at the beginning of this paper as the test area. Fig.~\ref{Fig3} displays the generated trajectories by all methods alongside corresponding heatmaps. Traditional random mobility models, i.e., RWP and GM, result in chaotic trajectories and heatmaps that bear no resemblance to the real-world patterns. While map-restricted models, M-RWP and M-GM, show some similarity in trajectories to real ones, they still fall short in distribution similarity due to the absence of a learning mechanism. As expected, DiffTraj demonstrates high similarity to real trajectories and heatmaps because it is trained directly on real data. Our proposed Map2Traj, even in a zero-shot scenario, produces results comparable to real trajectories and even surpasses DiffTraj in trajectory details. More experiment results are presented in the supplementary material.

\begin{table*}[!h]
    \centering
    \begin{tabular}{ccccc}
        \toprule \multirow{2.5}{*}{\textbf{Mobility model}}  & \multicolumn{2}{c}{\textbf{Trajectory similarity}} & \multicolumn{2}{c}{\textbf{Distribution similarity}}
        \\
        \cmidrule(r){2-3}\cmidrule(r){4-5}
        &EDR ($\tau=20$) $\downarrow$ & \ \ DTW $\downarrow$ \ & Cosine similarity $\uparrow$ & Wasserstein distance $\downarrow$\\
        \midrule 
        RWP & 264.1 & 76.74 & 0.1537 & 21.01\\
        GM &213.1& 82.96  & 0.1698 & 19.21\\
        M-RWP & 192.0 & 21.44 & 0.3081 & 22.91\\ 
        M-GM & 155.4 & 33.57  & 0.2793 & 26.32\\
        DiffTraj & 68.35 & 13.63 & 0.5573 & 9.134\\
        \textbf{Map2Traj} &\textbf{21.47} & \textbf{8.933}& \textbf{0.6834} & \textbf{6.096}\\
        \midrule 
        Real trajectories & 7.570 & 1.018 & 0.9959 &  2.569\\
        \bottomrule
    \end{tabular}
    \caption{Quantified evaluation of trajectory generation performance}
    \label{tab2}
\end{table*}

Further, we evaluate the similarity of the generated trajectories with real ones using the above-mentioned metrics. 
Considering the stochastic nature of trajectory generation, we generate 1,000 trajectories for each method and compare them against a benchmark of 1,000 real trajectories. For each generated trajectory, we calculate the similarity metric with all real trajectories, identifying the minimum value as the representative metric. This process is repeated for all the 1,000 trajectories to determine the average trajectory similarity between the generated set and the real dataset.

For distribution similarity, we aggregate the 1,000 trajectories into a 192$\times$192 binary image, normalizing them to represent the probability of user presence at various locations. We also calculate metrics for another set of 1,000 real trajectories to serve as optimal similarities. Table~\ref{tab2} presents the quantified similarity comparison among different trajectory sets. The results indicate that our proposed Map2Traj significantly exceeds the random mobility models and map-restricted ones, producing synthetic trajectories closely resembling real ones in both trajectory and distribution similarities. This suggests that Map2Traj has effectively learned the correlation between street maps and actual trajectories.

It is encouraging to see our zero-shot Map2Traj model outperforms the area-specific DiffTraj. The advantage of Map2Traj lies in the continuous guidance from street maps throughout the denoising process, while DiffTraj only involves some trip information as the condition. On the other hand, DiffTraj fixes the trajectories into uniform shapes through sampling, instead of transforming them into images, potentially leading to information loss. However, it is crucial to acknowledge that learning-based methods, such as DiffTraj, have the potential to outmatch Map2Traj given a sufficiently large training dataset, a wider and deeper network structure, or in some specific test scenarios. The primary contribution of this work is the development of a zero-shot trajectory generation method, rather than merely surpassing existing area-specific trajectory generation techniques.

\subsection{Case Study of Map2Traj in Wireless Network}
In this case study, we employ Map2Traj in wireless network optimization, specifically for the user association and load balancing tasks.

\subsubsection{System Model and Task Overview}
\begin{figure}[!h]
\centering
\includegraphics[width=0.45\textwidth]{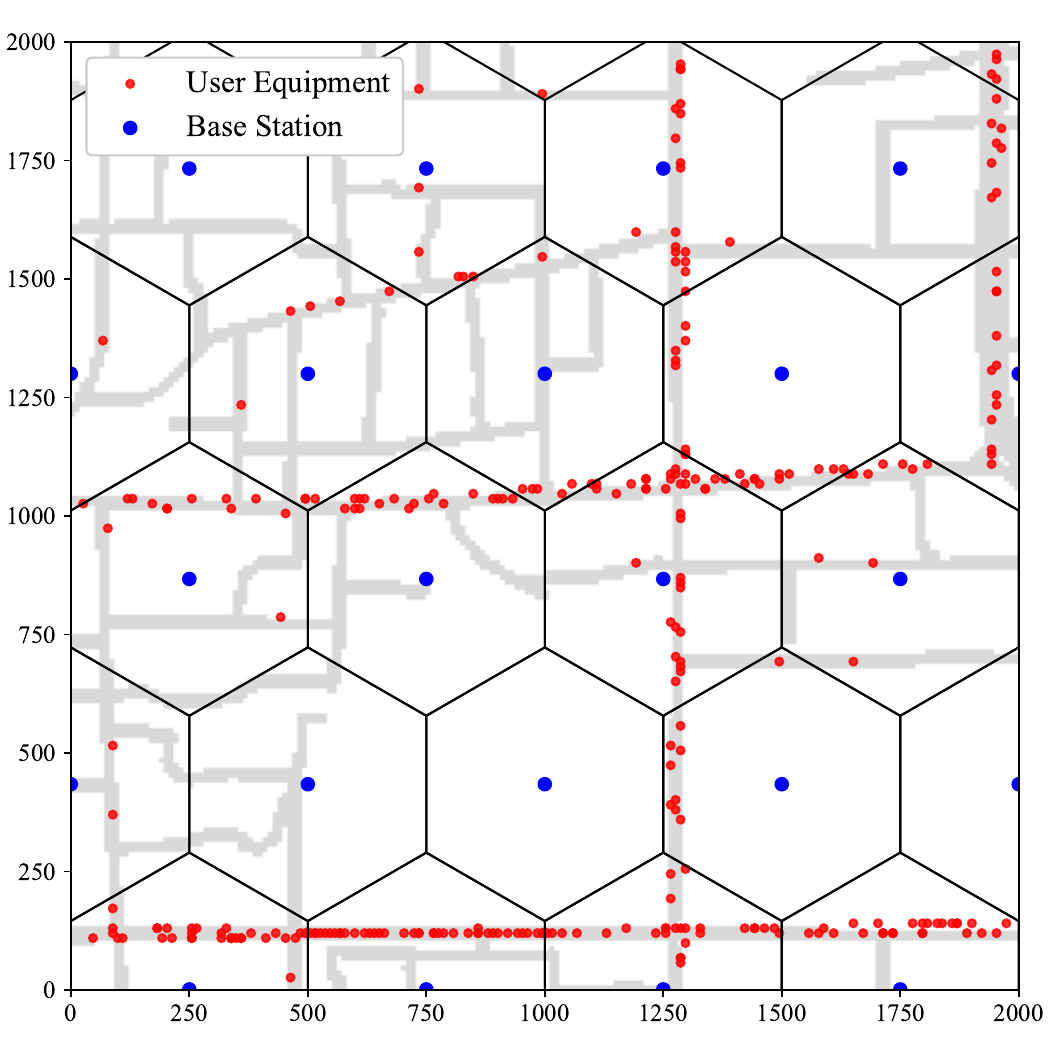}
\caption{Scenario of user association and load balancing.}
\label{Fig4}
\end{figure}

We consider a typical urban area, i.e., the test area in the last section, where base stations are densely deployed in a hexagonal pattern, with a 500 m interval. Each base station possesses multi-band capabilities, supporting connections at 3.7 GHz with a 40 MHz bandwidth and 0.7 GHz with a 10 MHz bandwidth. Users move continuously within this area, as illustrated in Fig.~\ref{Fig4}, connecting to base stations based on a user association policy. Traditional user association methods that maximize signal-to-interference-plus-noise ratio (SINR) can lead to load imbalances and frequent handovers, resulting in user rate degradation when users are not uniformly distributed. An advanced user association strategy is essential to balance network loads and minimize handovers, thereby enhancing user experience and connection stability.

\subsubsection{Mehthodology}

Even without considering the prediction problem related to handover, solving the integer linear programming problem for user association is unfeasible due to real-time requirements and computational complexity arising from numerous users. In this context, current works primarily employ DRL to address this problem. However, the direct training of DRL agents within the real-world wireless network is fraught with challenges, including prohibitive trial-and-error costs and the risk of compromising the quality of service \cite{9372298}. Consequently, constructing a realistic training environment is crucial for applying DRL in wireless network optimization. While there is extensive research on wireless channel measurement and modeling, studies on user mobility models are limited. The use of random mobility models, including RWP and GM, prevalent in current studies \cite{8796358,9329087,9685781}, does not ensure the efficacy of DRL agents when applied to real-world scenarios. To mitigate this, we propose the adoption of Map2Traj as the user mobility model in training environments to enhance the performance of DRL agents.


\subsubsection{Case Study Setup and Metrics}

The main purpose of this case study is to examine whether there is a performance degradation for DRL agents trained with traditional random mobility models and deployed in the real environment and whether the introduction of Map2Traj can alleviate such a phenomenon. The case study consists of two phases. Initially, DRL agents are trained in environments based on various mobility models, including RWP, GM, M-RWP, M-GM, and Map2Traj. After achieving convergence, these agents are deployed into the real environment where user movements adhere to real trajectories to assess performance. The performance of a DRL agent trained directly in the real environment is also provided as a benchmark for optimal performance. 

To focus on mobility model comparisons, wireless channels are kept constant across different environments, using the urban macrocell path-loss model from 3GPP \cite{3gpp} and the shadow fading model implemented via the sum-of-sinusoids method, as used in QuaDRiGa \cite{quadriga_website}.
The DRL method employed here is the state-of-the-art proximal policy optimization (PPO) algorithms \cite{schulman2017proximal}, with the actor-network built on a long short-term memory (LSTM) network \cite{hochreiter1997long} to incorporate memory capabilities.
Performance metrics encompass the 5th percentile user rate (5\% rate) to evaluate cell-edge performance, and the logarithmic mean of all user rates, which serves as an indicator of the overall utility of the wireless network.

\subsubsection{Results}
\begin{table}[]
    \centering
    \begin{tabular}{cccc}
        \toprule
        \textbf{Environment} & \textbf{Method} & \textbf{$5\%$ rate ($\times 10^5$)} $\uparrow$ & \textbf{Utility} $\uparrow$\\
        \midrule 
        \multirow{2}{*}{RWP} & Max SINR & 5.030 & 6.561  \\
        & DRL & 7.241 & 6.672 \\
        \midrule
        \multirow{2}{*}{GM} & Max SINR & 7.662 & 6.721\\
        & DRL & 8.881 & 6.788  \\
        \midrule
        \multirow{2}{*}{M-RWP} & Max SINR & 5.250 & 6.603  \\
        & DRL & 9.752 & 6.707 \\
        \midrule
        \multirow{2}{*}{M-GM} & Max SINR & 5.948 & 6.674\\
        & DRL & 11.373 & 6.766  \\
        \midrule
        \multirow{2}{*}{Map2Traj} & Max SINR & 5.871 & 6.599 \\
        & DRL & 9.877 & 6.705 \\
        \midrule
    \end{tabular}
    \caption{User association performance after training}
    \label{tab3}
\end{table}
We delineate the comparative performance of DRL agents across various training environments in Table~\ref{tab3}, and provide a GIF in the supplementary material to dynamically illustrate user movement across various mobility models. Upon achieving adequate convergence, it is observed that all DRL agents notably surpass the traditional Max SINR approach. Subsequently, these agents are deployed in the real scenario, and the results are presented in Table~\ref{tab4}. Consistent with our expectations, agents trained with random mobility models exhibit substantial performance degradation, in some cases deteriorating to levels comparable to or even worse than the Max SINR method. In contrast, the DRL agent trained with the Map2Traj-based mobility model maintains superior performance over the Max SINR method.

\begin{table}[]
    \centering
    \begin{tabular}{ccc}
        \toprule
        \textbf{Method} & \textbf{$5\%$ rate ($\times 10^5$)} $\uparrow$ & \textbf{Utility} $\uparrow$\\
        \midrule
         Max SINR & 5.408 & 6.575\\
         DRL (RWP) & 2.238 & 6.576 \\
         DRL (GM) & 2.771 &  6.587\\
         DRL (M-RWP) & 5.568 & 6.637 \\
         DRL (M-GM) & 3.145 &  6.601\\
         DRL (Map2Traj) & \textbf{7.823} & \textbf{6.669}\\
        \midrule
         DRL (Real) & 8.538 & 6.684 \\

        \bottomrule
    \end{tabular}
    \caption{User association performance in real environment}
    \label{tab4}
\end{table}

\begin{figure}[!t]
\centering
\includegraphics[width=0.5\textwidth]{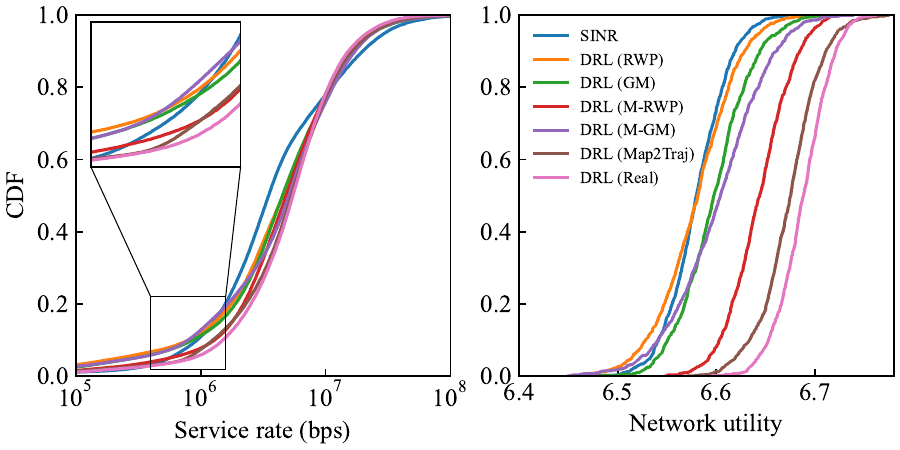}
\caption{CDF of performance in real environment.}
\label{Fig5}
\end{figure}

To provide a more nuanced view of the performance, we present the cumulative density function (CDF) of the metrics. As illustrated in Figure \ref{Fig5}, the Map2Traj-based DRL agent not only outperforms its counterparts trained with random mobility models but also closely approaches the performance of the agent trained in the real environment. All these results demonstrate that synthetic trajectories generated by Map2Traj have efficacy comparable to real ones for downstream applications like training DRL agents.




\section{Conclusion}
In this paper, we delve into the correlation between street maps and trajectories, introducing a novel Map2Traj method based on a diffusion model to achieve zero-shot trajectory generation. While prior research has successfully generated trajectories for specific regions using real datasets, we are the first to directly generate synthetic trajectories for new and unobserved areas. Extensive experiments demonstrate our method outperforms the traditional random mobility model and even the state-or-the-art area-specific model in terms of trajectory and distribution similarities with real ones. Furthermore, through a case study focused on wireless network optimization, we validate that trajectories generated by Map2Traj exhibit comparable efficacy to real ones for downstream applications. In future work, we intend to investigate additional applications of Map2Traj in the wireless communication realm and further enhance its performance and compatibility for areas of varying sizes.

\newpage

\bibliography{aaai25}

\appendix








\end{document}